\documentclass[12pt]{article}
\usepackage[sort&compress]{natbib}

\usepackage[utf8]{inputenc}
\usepackage{amsmath}
\usepackage{graphicx}
\usepackage{hyperref}
\usepackage{booktabs}
\usepackage{caption}
\usepackage{authblk}

\title{PolyIPA - Multilingual Phoneme-to-Grapheme Conversion Model}
\author[1]{Davor Lauc}  
\affil[1]{University of Zagreb, Faculty of Humanities and Social Sciences, Zagreb, Croatia\\
Email: \texttt{dlauc@ffzg.unizg.hr}, \texttt{davor@mondonomo.ai}}
\date{12/12/2024}     

\begin{document}

\maketitle

\begin{abstract}
This paper presents PolyIPA, a novel multilingual phoneme-to-grapheme conversion model designed for multilingual name transliteration, onomastic research, and information retrieval. The model leverages two helper models developed for data augmentation: IPA2vec for finding soundalikes across languages, and similarIPA for handling phonetic notation variations. Evaluated on a test set that spans multiple languages and writing systems, the model achieves a mean Character Error Rate of 0.055 and a character-level BLEU score of 0.914, with particularly strong performance on languages with shallow orthographies. The implementation of beam search further improves practical utility, with top-3 candidates reducing the effective error rate by 52.7\% (to CER: 0.026), demonstrating the model's effectiveness for cross-linguistic applications.
\end{abstract}

\section{Introduction}
Although text-to-speech and speech-to-text models nowadays are mostly end-to-end, there are still many use cases for intermediate representations using phonetic alphabets such as IPA. This is particularly evident in multilingual speech recognition systems, where IPA tokens enable the transfer of phonetic representations between different languages. In low-resource language scenarios, as little as 10 hours of target language data combined with IPA-based training from other languages can significantly reduce ASR error rates. A practical example of this approach can be seen in bilingual systems such as the Mandarin / Taiwanese speech recognition system, which achieved 92.55\% word accuracy using the Tong-yong Phonetic Alphabet (TYPA).

Recent advances in multilingual grapheme-to-phoneme (G2P) conversion have demonstrated significant progress through various neural architectures and approaches. The Transformer-based ByT5 model has shown remarkable success in handling approximately 100 languages simultaneously, offering improved phone error rates through joint learning and zero-shot prediction capabilities. Such byte-level representation models have proven particularly effective, achieving 16.2-50.2\% relative word error rate improvements over character-based counterparts while maintaining smaller model sizes for languages with diverse writing systems.

Neural transducer models utilizing explicit edit actions and trained with imitation learning have also demonstrated competitive performance, particularly when adapted to use substitution edits and trained with weighted finite-state transducers. The combination of multilingual Transformers with self-training ensembles has yielded impressive results across 15 languages, achieving a 14.99 word error rate and a 3.30 phoneme error rate. Furthermore, models incorporating global character vectors (GCVs) with bidirectional recurrent neural networks have shown promise, achieving over 97. 7\% syllable accuracy in Japanese, Korean, Thai and Chinese.

This research is focused on the reverse problem, phoneme-to-grapheme (P2G) conversion. Despite the well-established nature of grapheme-to-pho\-ne\-me (G2P) conversion in speech technologies, its reverse direction, phoneme-to-grapheme (P2G) conversion, has received significantly less research attention in the field. However, recent studies have demonstrated the value of P2G systems in various applications, particularly in improving end-to-end ASR systems and addressing the challenge of out-of-vocabulary words.

\section{Problem Statement and Contributions}
The main focus of this paper is the development of the P2G model that is usable in the use cases of transliteration and information retrieval, as well as onomastic research. The goal is to achieve approximately correct transcription of the phonetic encoding in the target language, assisting in multilingual transliterations and translation tasks, sound-like retrieval, and finding similarly sounding proper names across the world languages.

This paper leverages existing work on developing G2P datasets and models. The contributions, besides simply reversing the direction of the task, are extensive data augmentation by developing two helper models:
\begin{itemize}
    \item Embedding model IPA2vec used to find soundalikes across languages 
    \item Sequence-to-sequence model similarIPA used to cover phonetic notations variations across different training data source
\end{itemize}

An additional contribution is fine-tuning the model by extending the G2P data set with phonetic transliteration of proper names that are underrepresented in available dictionaries, even though proper names dominate vocabulary \citep{Lauc2024}.

\section{Data Collection and Preparation}
\subsection{Data Sources}

The largest dataset available, WikiPron represents a foundational multilingual resource containing 1.7 million pronunciations in 165 languages \citep{Lee2020}. The data is provided in IPA format with clear transcription guidelines for contributors. CharsiuG2P dataset offers a transformer-based solution that supports 100 languages, built on the ByT5 architecture \citep{Zhu2022}. The Montreal Forced Aligner Dictionary Collection provides extensive pronunciation data formatted specifically for speech alignment tasks \citep{McAuliffe2017}, but is useful for other tasks as well. PWESuite serves as an evaluation framework for phonetic word embeddings, offering tasks such as sound similarity correlation and cognate detection \citep{Zouhar2023}.

Based on analysis of language representation versus language sizes, several language-specific sources were also included. KaamelDict stands as the largest Persian G2P dictionary with over 120,000 entries \citep{Fetrat2024}, that unifies multiple phonetic representation systems and provides detailed phoneme-to-IPA mapping. Thai G2P \citep{Phatthiyaphaibun2020} focuses on Thai language specifics. Several other language-specific resources include: Rulex for Russian \citep{Poretsky2022}, IcePronDict for Icelandic \citep{Grammatek2023}, German IPA Dictionary \citep{Devio2023}.

\subsection{Data Cleaning and Normalization}
Data sources utilizing non-IPA transcription systems, such as X-SAMPA \citep{wells_computer-coding_2016} and ARPABET \citep{Klautau2001}, were converted to standard International Phonetic Alphabet (IPA) notation. Particular attention was paid to the accurate conversion of extended and language-specific symbols, including tonal markers. The language names and codes from all sources were standardized using ISO-639-1 language codes when available or ISO-639-3 codes when necessary. Datapoints with unidentified languages, comprising 0.03\% of the dataset, were excluded. Multiple phonetic transcriptions for a single grapheme within one language were retained.

The initial dataset comprised 20,987,451 unique language-grapheme-phoneme triples. All strings underwent NFC normalization, case lowering, and deduplication. To address invalid transliterations identified in the source data, we implemented a validation function to verify that the phonetic transcriptions contained only valid symbols of the extended IPA alphabet. This cleaning process resulted in 20,087,067 triples. Further refinement involved removing entries where the detected script did not correspond to the official script of the respective language, eliminating both romanized versions of non-Latin languages and various noise elements. The final data set contained 19,648,870 data points.

\subsection{Data Augmentation}
\subsubsection{Soundalikes by IPA2VEC model}

To address the identified gaps in phonetic pattern coverage across languages, we implemented a data augmentation approach based on phonetic similarity mining using a Siamese neural network architecture.

The training data set was constructed using the large-scale cognate lexical database \citep{Batsuren2019}, filtering for entries that existed in our phonetic dictionary. For each positive pair in the database, we generated one random negative example, resulting in 2,364,433 word pairs. Each pair was annotated with a feature edit distance score following the methodology proposed by \citet{Mortensen2016}. The dataset was split into training, development, and test sets in a 98:1:1 ratio. 

To generate phonetic embeddings, we implemented a Siamese neural network architecture utilizing ByT5 \cite{Xue2022} as the shared encoder. The model was trained using cosine similarity loss with mean pooling of token embeddings over 30 epochs, achieving final training and evaluation losses of 0.00075 and 0.00072 respectively.

The data augmentation process employed a two-stage similarity search approach:

1. Initial candidate generation using approximate nearest neighbor search via FAISS (Johnson et al., 2019), retrieving 10,000 closest matches per entry

2. Refinement through precise feature edit distance calculation \cite{Mortensen2016}, with a distance threshold of 5.

Applied to the training dataset, this process generated 1.1 million phonetically similar pairs, from which ones where variants already existed in the original sources were eliminated. The number of examples generated by this method was 927,351.

\subsubsection{SimilarIPA model}
The second data augmentation approach addressed the variations in IPA transcription practices between different sources and annotation traditions. These variations are particularly evident when combining multiple phonetic data sets with different transcription conventions.

We generated additional training data by leveraging cases where multiple IPA transcriptions existed for the same language-grapheme pair in our dataset. For each language-grapheme pair with multiple valid IPA transcriptions, we extracted all possible combinations to create IPA-to-IPA transformation pairs. This process yielded 4.4 million pairs, which were split into training sets (98\%), evaluation sets (1\%) and test sets (1\%).

For modelling these transformations, we employed the T5.1 architecture (Raffel et al., 2020), which has demonstrated strong performance in sequence-to-sequence tasks. We first trained a custom SentencePiece vocabulary of 32K tokens \cite{KudoRichardson2018}, on the training data to get vocabulary optimized for phonetic character sequences. The T5.1 small model was then trained from scratch for five epochs.

The model achieved an evaluation loss of 0.21 and a Character Error Rate (CER) of 0.06. These metrics are considered satisfactory for the purpose, especially given the inherent many-to-many nature of the task --- when a grapheme has more than two valid IPA transcriptions, each input can correspond to multiple valid outputs.

The trained model was then applied to generate alternative IPA transcriptions for the entire dataset. For each IPA transcription in our corpus, we generated up to three alternative representations, filtering out any generations with a character error rate (CER) greater than 0.15 compared to the input to ensure quality. Variants that already existed in collected sources were excluded. The total number of examples generated in this way was 5,526,352. 

These alternative transcriptions and soundalikes generated by the IPA2VEC model helped bridge the gaps between different transcription conventions and provided additional training data that captured legitimate variations in how the same sounds can be represented in IPA. This augmentation proved particularly valuable for languages where the original dataset contained relatively rigid or source-specific transcription patterns, helping the final P2G model to become more robust to input variations. 

In the future research, this augmentation process should be improved firstly by developing better metrics for phonetic similarity. Although the issue of measuring the perception of sound similarity is highly subjective and controversial in the phonetic community \cite{Hazan2017}, it seems that if the language background of the subject is taken into consideration it could be improved \cite{Antoniou2023}. So we hypothesize that the development of a unified model that would, for a given IPA string and language, generate IPA rendering of similar sounding words for a typical speaker of a given language would significantly improve the data augmentation process and the quality of the final model.

\subsection{Accent and Tone Reduction}
The final augmentation process included careful handling of the IPA transcriptions through multiple normalization steps. First, a validation function was implemented to ensure the quality of phonetic transcriptions. This function performed several checks:

\begin{itemize}
    \item Basic validation of input type and format
    \item Removal of diacritical marks and tone indicators using Unicode normalization
    \item Filtering of special symbols while preserving valid IPA characters through regex pattern matching
    \item Verification using the ipapy library's IPA validation
\end{itemize}

This cleaning was particularly important for handling variations in transcription practices across different data sources, where the same sound might be annotated with different diacritical marks or tone indicators. After this normalization, the data set retained the core phonetic information while reducing noise from inconsistent transcription practices.

\subsection{Language and Script Handling}

Data preparation included special handling for languages that use multiple scripts. Languages like Chinese (\texttt{zh}) and Serbian (\texttt{sr}), which can be written in different scripts, were encoded with script-specific language codes (e.g., \texttt{ zh\_Hani} for Chinese in Han characters, \texttt{ sr\_Latn} for romanized Serbian). This distinction was crucial for training the model to handle script-specific transcription patterns. Each IPA transcription was prefixed with a language-script code in the format \texttt{<\{lang\_code\}>} or \texttt{<\{lang\_code\}\_\{script\}>} when necessary.

\subsection{Dataset Splitting and Upsampling}

The dataset was split into training and test sets using a stratified sampling approach to ensure representative coverage across languages. The key characteristics of the splitting process included:

\begin{itemize}
    \item A fixed test and evaluation set size of 5,000 examples
    \item Stratification by ISO language code to maintain language distribution
    \item Balanced sampling to prevent over-representation of high-resource languages
\end{itemize}

The upsampling strategy was integrated directly into the training data generation process, with several key components working in concert. First, a length filter was applied to ensure that all examples remained under 40 tokens, maintaining computational efficiency and model stability. During generation, the system produced multiple variants of each training example:
\begin{enumerate}
    \item added clean versions of IPA transcriptions by removing special symbols and diacritics if it is different from the original
    \item incorporated phonetically similar transcriptions identified by the similarity models
    \item repeating original examples to maintain proper distribution balance
\end{enumerate}

This dynamic generation approach allowed flexible data augmentation while avoiding explicit storage of the augmented data set. The process checked for duplicates against an index of existing training data to prevent redundancy, effectively increasing the diversity of training examples while preserving the natural distribution of phonetic patterns across languages.

The final size of the training dataset contained 79,320,217 examples, which were pretokenized and saved to a parquet file for efficient training.

\section{Experimental Setup}
The model was trained using the Seq2SeqTrainer from the Hugging Face Transformers library, with optimized training configurations for the P2G task. The core configuration included:

\begin{itemize}
    \item Model Architecture: Two variants were tested - ByT5-small and MT5-small \citep{Xue2022}, with ByT5 showing superior performance for byte-level processing of multilingual text
    \item Maximum Sequence Length: 64 tokens
    \item Learning Rate: 4e-5 with a linear warm-up over 1000 steps
    \item Weight Decay: 0.01 for regularization
    \item Batch Size: 96 per device with gradient accumulation steps of 4
    \item Training Duration: 3 epochs
    \item Gradient Clipping: Maximum gradient norm of 1.0 to prevent exploding gradients
\end{itemize}

The training used a custom DataCollator for efficient batch processing and a ParquetIterableDataset for memory-efficient handling of the large augmented data set (79.3M examples). Training was tested with both ByT5-small and MT5-small models. As the initial validation metrics of ByT5 were improving much more quickly than those of the small MT5 model, corroborating research by \citet{Lee2020}, only training of the ByT5 model was continued for three epochs. It should be noted that, due to the repetition of original examples from the IPA dictionary in train DS, as described before, it would correspond to approximately 10 epochs without repetition.

The training metrics progressed steadily, with:

\begin{itemize}
    \item Training loss decreasing from 4.13 to 0.075 over three epochs
    \item Evaluation loss following a similar trajectory, dropping from 3.9413 to $\sim$0.073
    \item Character Error Rate (CER) improving from 0.9531 to 0.039
\end{itemize}

Although the obtained evaluation metrics are satisfactory, especially considering that the training set contains multiple output sequences for the same language-phoneme combination, an initial informal evaluation showed that in some cases the model learned grapheme representations of phonemes from other languages, as the most probable output sequence did not reflect the phonetics of the target language but rather existing examples from other languages. A typical example is the rendering of the voiced labiodental fricative ``v'' in German. Although it should be rendered as ``w'', sometimes the most probable sequence was rendered using ``v'', which represents the sound ``f''. The proper outputs sometimes appeared in the second or later positions of the most probable sequences. To account for this, an additional augmentation of the data set was performed.

\subsection{Fine-tuning Model for Proper Names}

The augmentation process began with a comprehensive phonetic analysis of common proper names across world languages. The initial data set consisted of the one million most frequent proper names from global language sources, as collected by Mondonomo Nomograph DB \citep{Lauc2024}. These names were processed through the trained phoneme-to-grapheme model, each token for 20 randomly selected languages, using a random function weighted by language size. For each input, the model generated 30 candidate sequences using a beam search with 90 beams (3 times the number of requested outputs) and early stopping to ensure diverse yet high-quality predictions. This process created a comprehensive dictionary mapping each language-grapheme combination to its possible IPA realizations.

The generated outputs were then analysed using a feature-based distance metric from the panphon library, which evaluates phonetic similarity based on articulatory features. For each generated sequence, the system calculated its phonetic distance from the original input and retained only one alternative demonstrating high phonetic similarity (distance less than 1\% of the maximum possible feature difference). The process utilized parallel processing with chunked data processing (1 million rows per chunk) to efficiently handle the large volume of comparisons. Special care was taken to normalize the IPA strings by removing diacritical marks and special symbols before distance calculation to ensure consistent comparison. The distance calculation was implemented using the Jandrec-Traversky feature edit distance algorithm, which provides a more nuanced measure of phonetic similarity compared to simple character-based metrics. The total number of new examples generated by this method was 482,475.

This data-driven approach served two purposes:
\begin{enumerate}
    \item It identified phonetically plausible alternative transcriptions for common proper names, effectively expanding the model's understanding of cross-linguistic name adaptation patterns
    \item It helped validate and filter the model's non-primary predictions, ensuring that alternative suggestions maintained phonetic coherence within each target language's sound system
\end{enumerate}

By focusing on proper names, which often exhibit unique phonetic adaptation patterns when moving between languages, this augmentation strategy enhanced the model's capability to handle the specific challenges of name transliteration and cross-linguistic adaptation.

These data were added to the existing training dataset, shuffled, and the model was trained for two additional epochs with the new dataset. The evaluation and test data sets were kept the same. The evaluation metrics after training (on the unchanged evaluation dataset) after the two epochs were 0.063 evaluation loss and 0.031 CER. The improvements could be attributed either to the extended data set or to the extended training or both, but the additional data definitively did not hinder the quality of previous training, as measured by the initial evaluation data set\footnote{The latest model is published to HF PolyIPA) and deployed to \href{nelma.mondonomo.ai}{https://nelma.mondonomo.ai} integrated with the IPA dictionary for common names.}.

\section{Results and Discussion}

The evaluation methodology employs multiple complementary metrics to assess the model's performance in phoneme-to-grapheme conversion across different languages. The primary metrics include Character Error Rate (CER), Character Level BLEU Score, and Top N Word Error Rate (WER), providing a comprehensive assessment of the model's accuracy and generation quality.

The character error rate (CER), calculated using the Levenshtein distance normalized by the reference length \cite{Levenshtein1966}, serves as our primary metric for measuring transcription accuracy. This metric is particularly suitable for phoneme-to-grapheme conversion tasks, as it operates at the character level and captures insertion, deletion, and substitution errors \cite{Niu2019}. Additionally, we compute character-level BLEU scores using SacreBLEU \cite{Post2018}, treating individual characters as tokens to evaluate the model's ability to preserve character sequences in the transliteration process. This approach has been shown to be effective in previous transliteration studies \cite{Najafi2018}.

To evaluate the performance of the model with beam search generation, we implement a top-N evaluation strategy similar to that used in machine translation studies \cite{Wu2016}. For each input sequence, the model generates N candidate outputs (N=1,3,5), and we calculate the WER for each candidate, recording both the best score and its position on the beam. This approach helps assess the model's ability to generate multiple valid transliterations, which is particularly important for languages with multiple acceptable grapheme representations for the same phoneme sequence.

The evaluation is stratified by language to account for the varying complexity of phoneme-to-grapheme mapping across different writing systems. For each language, we calculate the mean performance metrics along with their standard deviations, weighted by the number of test samples. This stratification helps identify language-specific patterns and potential biases in the performance of the model, similar to the approach used in the evaluation of multilingual NLP \cite{Hu2020}.

\subsection{Overall Performance}
The model demonstrates strong performance in the majority languages, with:
\begin{itemize}
    \item Mean Character Error Rate (CER): 0.055 ($\pm$0.167)
    \item Mean Character-level BLEU: 0.914 ($\pm$0.212)
    \item Exact Match Accuracy: 0.830 ($\pm$0.376)
    \item Top-3 WER: 0.026 ($\pm$0.115)
\end{itemize}

\begin{figure*}[t]
    \centering
    \includegraphics[width=\textwidth]{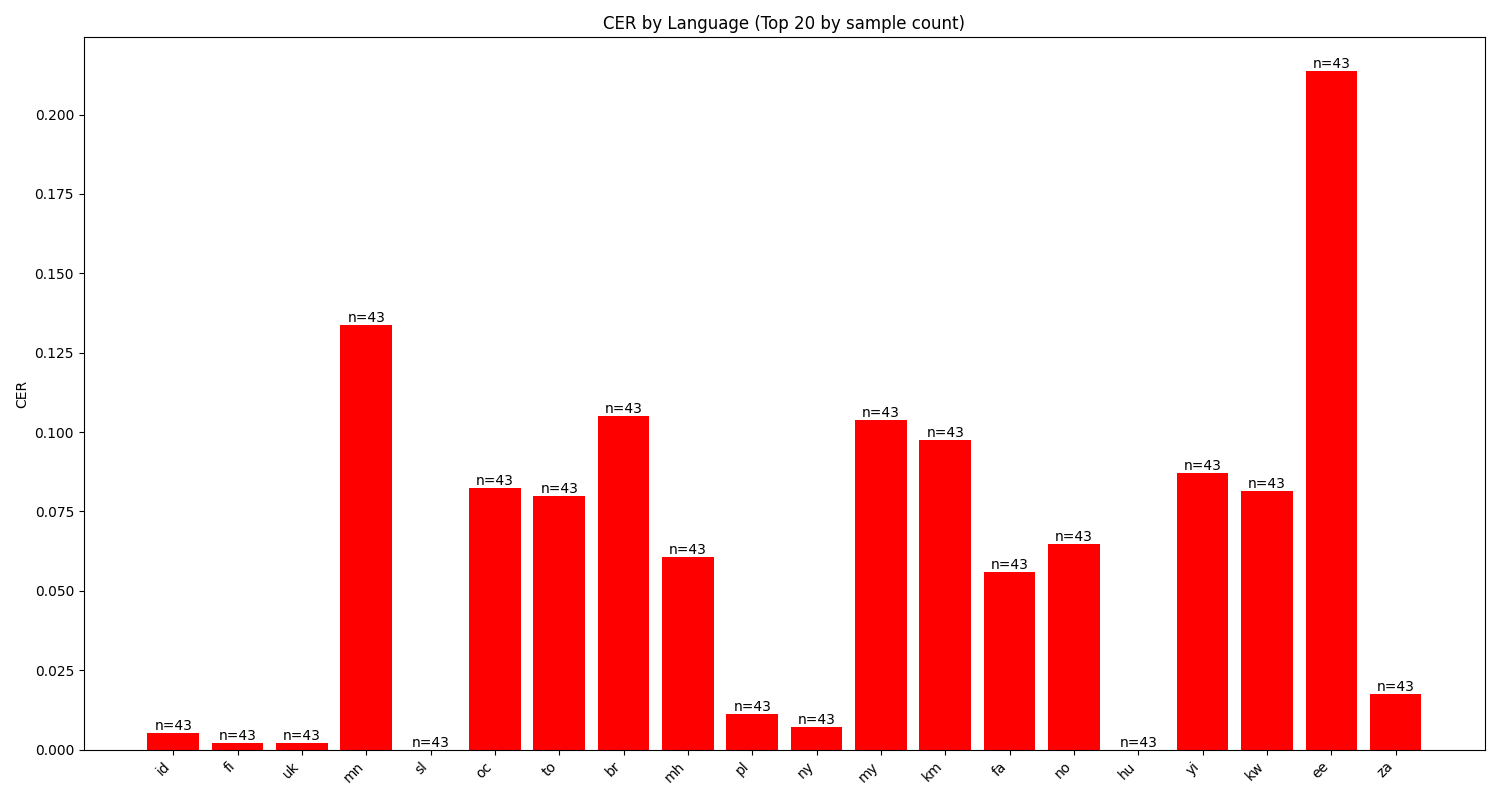}
    \caption{Distribution of Character Error Rates across all languages}
    \label{fig:cer_distribution}
\end{figure*}

The high exact match rate and low CER indicate that the model successfully learns phoneme-to-grapheme mappings for most languages.

\subsection{Orthographic Depth Analysis}
Performance correlates strongly with orthographic depth, a concept developed by \citet{Katz1992} in their Orthographic Depth Hypothesis. Languages can be classified along a continuum from shallow (transparent) to deep (opaque) orthographies:

\subsubsection{Shallow Orthographies (High Performance)}
\begin{itemize}
    \item Finnish (CER: 0.002): Known for nearly perfect grapheme-phoneme correspondence
    \item Spanish (CER: 0.000): Highly regular orthography
    \item Turkish (CER: 0.000): Consistent letter-sound relationships
    \item Croatian (CER: 0.000): Transparent orthographic system
\end{itemize}

\subsubsection{Deep Orthographies (More Challenging)}
\begin{itemize}
    \item English (CER: 0.067): Complex historical influences \citep{Venezky1970}
    \item French (CER: 0.120): Multiple grapheme-phoneme mappings
    \item Chinese (CER: 0.522): Morphosyllabic writing system
\end{itemize}

\subsection{Language Family Performance}
\subsubsection{High-Performing Language Families}
\begin{enumerate}
    \item Uralic Family:
    \begin{itemize}
        \item Hungarian (CER: 0.000)
        \item Finnish (CER: 0.002)
    \end{itemize}
    Reflects Abercrombie's (1967) observation about phonemic writing systems

    \item Turkic Family:
    \begin{itemize}
        \item Turkish (CER: 0.000)
        \item Uzbek (CER: 0.019)
    \end{itemize}
    Demonstrates success with agglutinative languages
\end{enumerate}

\subsubsection{Challenging Language Families}
\begin{enumerate}
    \item Sino-Tibetan:
    \begin{itemize}
        \item Chinese (CER: 0.522)
    \end{itemize}
    Consistent with Share's (2008) analysis of non-alphabetic writing systems

    \item Celtic:
    \begin{itemize}
        \item Manx (CER: 0.296)
        \item Scottish Gaelic (CER: 0.115)
    \end{itemize}
    Reflects historical orthographic complexity noted by \citet{Sproat2000}
\end{enumerate}

\subsection{Script Analysis}
Performance patterns align with \citet{PerfettiLiu2005}'s Universal Writing System Constraint:
\begin{enumerate}
    \item Latin script languages: Strong performance
    \item Syllabic scripts (Thai, Khmer): Moderate performance
    \item Logographic scripts: Lower performance
    \item Alphabetic non-Latin: Good performance
\end{enumerate}

\subsection{Beam Search Analysis}

The beam search results demonstrate:

\begin{figure*}[t]
    \centering
    \includegraphics[width=\textwidth]{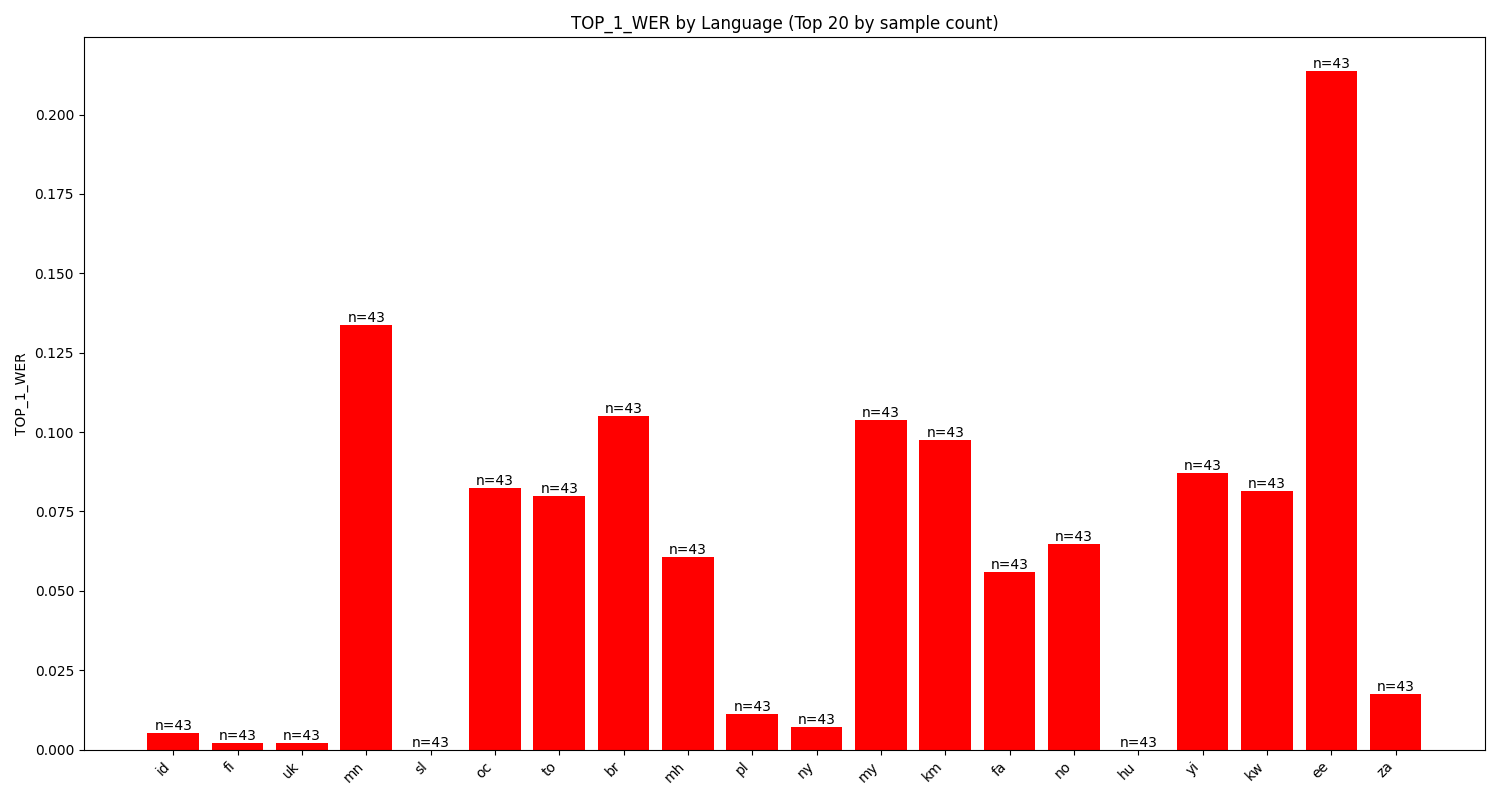}
    \caption{Top-1 Performance Comparison (Top 20 languages by sample count))}
    \label{fig:cer_top1}
\end{figure*}

\begin{figure*}[t]
    \centering
    \includegraphics[width=\textwidth]{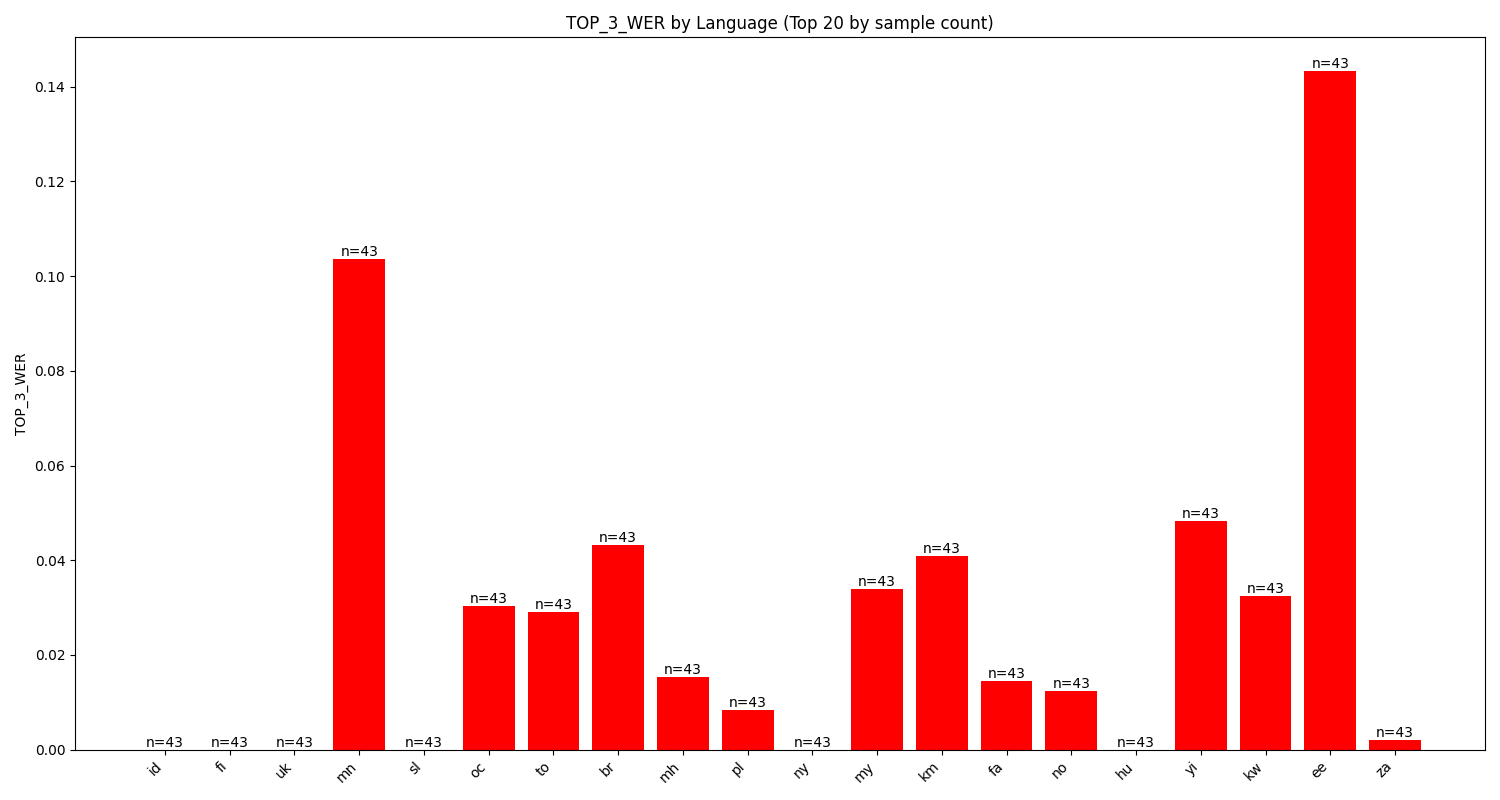}
    \caption{Top-3 Performance Comparison (Top 20 languages by sample count))}
    \label{fig:cer_top3}
\end{figure*}

\begin{enumerate}
\item Top-1 and Top-3 performance differences
\item Language-specific benefits from larger beam sizes
\item Correlation with orthographic depth
\end{enumerate}

\subsection{Error Pattern Analysis}
Common error patterns align with the \cite{Wiese2004} typology of orthographic complexity:
\begin{enumerate}
    \item Character substitutions in phonetically similar sounds
    \item  Diacritic mark variations
    \item Higher error rates in:
    \begin{enumerate}
        \item Non-phonemic orthographies
        \item Complex writing systems
        \item Limited training data case
    \end{enumerate}
\end{enumerate}

\section{Discussion}

The experimental results demonstrate both the capabilities and limitations of our multilingual phoneme-to-grapheme conversion approach. The model achieves strong performance across most languages, particularly those with shallow orthographies, while revealing systematic challenges that align with theoretical expectations from linguistic research.

\subsection{Performance Analysis}

The performance of the model exhibits clear patterns correlated with the linguistic characteristics of the target languages. Languages with shallow orthographies consistently show excellent results (CER < 0.005), supporting the Orthographic Depth Hypothesis \cite{KatzFrost1992}. Finnish, Spanish, Turkish, and Croatian languages, all languages with highly regular grapheme-phoneme correspondence, achieve near-perfect accuracy. This aligns with previous findings in cross-linguistic studies of writing systems \cite{Share2008}.

However, significant challenges arise with deeper orthographies. English and French show moderate error rates (CER of 0.067 and 0.120 respectively), reflecting their complex historical influences on spelling conventions. The most challenging cases appear in logographic writing systems, particularly Chinese (CER: 0.522), where the relationship between phonemes and graphemes is fundamentally different from alphabetic systems.

\subsection{Error Analysis}
Detailed error analysis reveals several systematic patterns:

\begin{enumerate}
    \item {\bf Phonetic Proximity Errors}: The most common substitution errors occur between phonetically similar sounds, particularly within the same manner or place of articulation. This suggests that the model has learned meaningful phonetic relationships, but sometimes struggles with fine-grained distinctions.
    \item {\bf Script-Specific Challenges}: Performance varies significantly between writing systems, with Latin-based scripts showing the highest accuracy, followed by other alphabetic systems, syllabaries, and finally logographic scripts. This hierarchy aligns with Perfetti and Liu's \cite{PerfettiLiu2005} Universal Writing System Constraint.
    \item {\bf Resource Effects}: Languages with limited training data show a higher variance in performance, indicating the importance of quantity and quality of data in model training.
    
\end{enumerate}

\subsection{Beam Search Analysis}

The implementation of beam search significantly improves the model's practical utility. Although the accuracy of the top-1 provides strong results for many languages (mean CER: 0.055), including the top-3 candidates reduces the effective error rate by 52 7\% (to CER: 0.026). This improvement is particularly pronounced in languages with multiple valid grapheme representations for the same phoneme sequence.

\section{Limitations}
Several limitations warrant acknowledgment:

\begin{enumerate}
    \item The model's performance degrades significantly for languages with complex morphophonological rules that are not captured in the training data.
    \item The current approach does not explicitly handle tone languages, treating tonal markers as diacritics rather than integral phonological features.
    \item The model occasionally produces phonologically valid but historically incorrect transliterations for proper names, particularly when crossing language families.

\end{enumerate}

\section{Future Work}
Based on our findings and analysis, we identify several promising directions for future research.

\subsection{Model Architecture Improvements}
\begin{enumerate}
    \item Phonetic Similarity Models: Development of a unified model that would, for a given IPA string and language, generate IPA rendering of similar sounding words for a typical speaker of a given language.

    \item Script-Specific Encoders: Developing specialized encoding layers for different writing systems could better capture script-specific features and improve performance across various orthographies.
    
    \item Morphological integration: Incorporating morphological analysis could help handle complex derivational and inflectional patterns that affect grapheme selection.
    
\end{enumerate}

\subsection{Data Enhancement}
\begin{enumerate}
    \item \textbf{Targeted Data Collection}: Expanding the training data for underrepresented languages and scripts, particularly focusing on languages with complex orthographic systems.
    
    \item \textbf{Generating synthetic data}: Expanding the training data using well performing .

    \item \textbf{Comprehensive Data Cleaning}: Implementation of more thorough cleaning procedures to address errors primarily arising from Wiktionary parsing issues.

    \item \textbf{Historical Pattern Mining}: Incorporating historical linguistics data to better handle etymology-based spelling patterns, especially for proper names and borrowed words.
    
\end{enumerate}

\subsection{Feature Engineering}
\begin{enumerate}
    \item \textbf{Phonological Feature Integration:} Explicitly modelling phonological features could improve the handling of sound changes across language boundaries.
    
    \item \textbf{Prosodic Modelling:} Developing better representations for suprasegmental features such as tone and stress.
    
    \item \textbf{Enhanced Phonetic Similarity Metrics:} Improving metrics for measuring phonetic similarity by incorporating language-specific perceptual factors, as suggested by recent research on cross-linguistic phonetic perception \citep{Antoniou2023}.
\end{enumerate}

\subsection{Evaluation Framework}
\begin{enumerate}
    \item \textbf{Language-Specific Metrics:} Developing evaluation metrics that account for systematic differences in writing systems and acceptable variation in transliteration.
    
    \item \textbf{User Studies:} Conducting human evaluation studies to assess the practical utility of the model's outputs in real-world applications.
    
    \item \textbf{Cross-Linguistic Validation:} Expanding evaluation to include more systematic testing of cross-linguistic name adaptation patterns.
\end{enumerate}

\section{Acknowledgments}
Training of the model was supported by a grant given to Mondonomo by the Nvidia Inception program.

\bibliographystyle{apalike}
\bibliography{PolyIPA_arxiv_20241212}

\end{document}